\documentclass[a4paper,runningheads]{llncs}
\usepackage[latin9]{inputenc}
\usepackage{color}
\usepackage{url}
\usepackage{amssymb}
\usepackage{graphicx}

\makeatletter

\pdfpageheight\paperheight
\pdfpagewidth\paperwidth

\providecommand{\tabularnewline}{\\}

\usepackage{url}
\usepackage{wrapfig}
\usepackage{colortbl}
\definecolor{Green}{rgb}{0,0.88,0}
\definecolor{Red}{rgb}{0.88,0,0}
\definecolor{Yellow}{rgb}{0.88,0.88,0}
\definecolor{Wine}{rgb}{0.612, 0.192, 0.388}
\urldef{\mailsa}\path|{mastmeyer, fortmeier, handels}@imi.uni-luebeck.de|
\urldef{\mailsb}\path|{pieper, sw, tkapur}@bwh.harvard.edu|
\urldef{\mailsc}\path|{pieper, sw, tkapur}@bwh.harvard.edu| 

\newcommand{\keywords}[1]{\par\addvspace\baselineskip
\noindent\keywordname\enspace\ignorespaces#1}

\makeatother

\begin{document}
\mainmatter 


\title{Model-based Catheter Segmentation in MRI-images}

\titlerunning{Model-based Catheter Segmentation}

\author{{Andre Mastmeyer}\inst{1}, {Guillaume Pernelle}\inst{2}, {Lauren Barber}\inst{3}, {Steve Pieper}\inst{3}, {Dirk Fortmeier}\inst{1}, {Sandy Wells}\inst{3}, {Heinz Handels}\inst{1}, {Tina Kapur}\inst{3}}

\authorrunning{{Andre Mastmeyer}, {Guillaume Pernelle}, {Tina Kapur}}



\institute{University of L\"ubeck, Inst. of Medical Informatics
\and Imperial College London, Computational Neuroscience Lab
\and Brigham \& Women's Hospital and Harvard Medical School, Surgical Planning Lab} 

\maketitle


\toctitle{{*}{*}{*}}

\tocauthor{{*}{*}{*}}
\begin{abstract}
Accurate and reliable segmentation of catheters in MR-gui\-ded 
interventions remains a challenge, and a step of critical importance in clinical workflows. In this work, under reasonable assumptions,
mechanical model based heuristics guide the segmentation process 
allows correct catheter identification rates greater than 98\% (error$\leqslant$2.88
mm), and reduction in outliers to one-fourth compared to the state of the art.
Given distal tips, searching towards the proximal ends of the catheters
is guided by mechanical models that are estimated on a per-catheter
basis. Their bending characteristics are used to constrain the image
feature based candidate points. The final catheter trajectories are
hybrid sequences of individual points, each derived from model and
image features.
We evaluate the method on a database of 10 patient MRI scans
including 101 manually segmented catheters. 
The mean errors
were 1.40 mm and the median errors
were 1.05 mm. The number of outliers deviating more than
2 mm from the gold standard is 7, and the number of outliers deviating more than
3 mm from the gold standard is just 2.

\end{abstract}

\section{Introduction}


MRI imaging is a standard component in the tools for diagnosis, biopsy,
and treatment of numerous types of cancers \cite{jolesz2014IGTBook}
due to the superior soft tissue contrast it provides. While MRI scans
provide the physician with improved imaging of the tumor and adjacent
soft tissue (compared to x-ray, ultrasound, or CT), the artifacts
created in these scans by typical catheters
are much more difficult to interpret. For instance, artifacts created
by catheters with metalic stylets in x-ray or CT scans acquired during pelvic
brachytherapy are very distinct and amenable to automatic segmentation
using standard image-processing techniques in commercial products
such as Oncentra Brachytherapy\footnote{Elekta, Stockholm, Sweden}
and Eclipse\footnote{Varian Medical Systems, CA, USA}. Today, no
corresponding specialized automatic solutions exist for MRI \cite{song12,dimaio05}.


We believe that in the absence of reliable automatic segmentation
tools for this task, MRI-guided catheter-based interventions will not
gain mainstream acceptance; manual accomplishment of this task is
time consuming, tedious, and error-prone, and requires significant
concentration of the operator.
We propose a new highly accurate catheter segmentation method 
based on a search cone \cite{pernelle13}
and use a benchmark database with a significant
number of catheters using different MRI imaging protocols for validation.

\section{Related Work}

The appearance of catheters in MRI images is influenced by the imaging
parameters, susceptibility effects, and the direction of the static
magnetic field relative to that of the catheter \cite{song12,dimaio05}.
Another challenge is posed by varying, deflected catheters caused
by different forces interacting with them during insertion into the
body. Vesselness measures \cite{drechsler10} try to enhance tubular
structures such as vessels in the images, but suffer heavily from
oversegmentation. In \cite{okazawa06} a catheter segmentation
method for 2D B-mode ultrasound images using Hough transforms is proposed. For 3D MRI images in \cite{pernelle13}, a sequence of Fibonacci-sized conical search
regions is used. In \cite{goksel09} three catheter models are
compared in terms of deflection accuracy measured against a physical
experiment.

Here, we based our model based segmentation method on two assumptions:
(1) catheters for biopsy and treatment are typically introduced
from a firm base (known as template, fixture, or guide); (2) the overall
catheter deflection can be approximately modelled by a main force acting
on the catheter tip.

\section{Methods}

\textbf{Angular Spring Models} In \cite{goksel09} the most effective
model for the simulation of catheter bending was reported to be the
angular spring model. In this approach the catheter is regarded as
an array of a finite number of stiff line segments (cantilever rods)
connected by angular springs. Following these ideas and simplifying
the approach by neglecting model convergence over time for efficiency
reasons, we can use the following forward and backward schemes to
calculate segment-wise deflection angles $\alpha_{i}$, the total
deflection angle $\alpha_{i}^{sum}$ and segment orthogonal end point
forces $F_{i}$ in 2D. The calculation works either in a forward manner
starting from the fixed end (the proximal end of the catheter in our
case) or in a backward fashion starting from a given catheter tip
for a number of segments $N_{seg}$ ($i=0..N_{seg}-1$):

\begin{center}
\begin{tabular}{ccccc}
\hline 
 & Forward  & ~~~~~~  & Backward  & \tabularnewline
\hline 
 & $\alpha_{0}=0;\,F_{0}=F_{0}^{giv}=given$  &  & $\alpha_{0}^{sum}=est.;\,F_{0}=F_{0}^{est}=est.$  & \tabularnewline
\hline 
 & $\alpha_{i+1}=F_{i}/k_{a}$  &  & $F_{i+1}=\frac{F_{i}}{cos\left(\alpha_{i}^{sum}\right)}$  & \tabularnewline
 & $\alpha_{i+1}^{sum}=\sum_{j=0}^{i+1}\alpha_{i}$  &  & $\alpha_{i+1}=\frac{F_{i+1}}{k_{a}}$  & \tabularnewline
 & $F_{i+1}=F_{i}\cdot cos\left(\alpha_{i+1}^{sum}\right)$  &  & $\alpha_{i+1}^{sum}=\alpha_{0}^{sum}-\sum_{j=0}^{i+1}\alpha_{j}$  & \tabularnewline
\end{tabular}
\begin{equation}
\label{eq:Model-iter}
\end{equation}

\par\end{center}

where $k_{a}$ is the physical angular spring stiffness {[}$N/m${]}
of the catheter and $F_{0}^{giv}$ is the effective main force acting
on the catheter at the tip. The angle $\alpha_{i}^{sum}$ describes
the total segment angle compared to a non-bent catheter (reference
catheter) in the forward and backward simulation. It is easy to see
that in the backward simulation starting from the given catheter tip,
we need estimates for $\alpha_{0}^{sum}$ and orthogonal segment force
$F_{0}^{est}$.

\begin{figure}
\includegraphics[bb=120bp 0bp 1410bp 441bp,clip,scale=0.27]{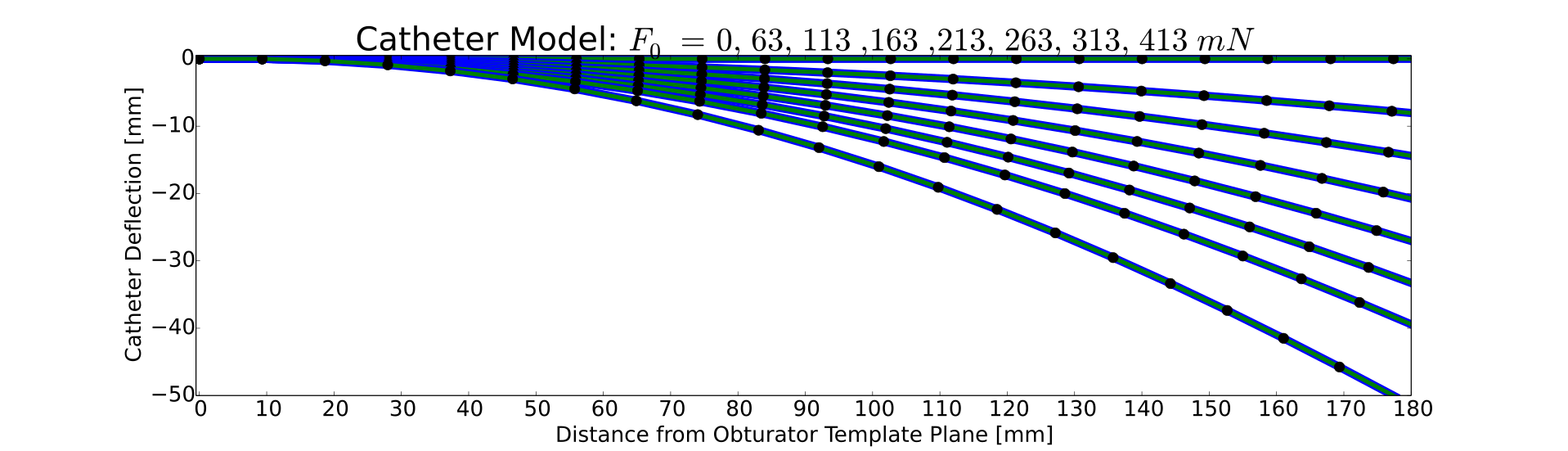}
\caption{\label{fig:Model-catheters}Model catheters and support points as
calculated by simulation. With estimated deflection $d$ and catheter
length $a$ as distance from the base template, we can identify the
most probable catheter and its parameters, e.g. $F_{0}$. For the
model look up table, in the gaps, linear interpolation is used.}
\end{figure}

\noindent \textbf{Model Estimation} The first step in our new algorithm
is the estimation of the angular spring model. For this aim, it is
necessary to have a reference catheter direction $\vec{r}$ representing
a non-deflected catheter. For many catheter grid based procedures,
the reference direction can be inferred using the base (or template)
that defines a normal plane $P$ which is assumed orthogonal to a
straight catheter\footnote{The estimation of the normal plane for specific clinical applications
can be a topic of investigation on its own. In the case of pelvic
brachytherapy with a template grid, four vitamin E capsules implanted
in the template create hyperintense artificats in the MR and specify
a regression plane for the reference catheter direction.}. Given the reference catheter vector $\vec{r}\bot P$ with a 
declination $\alpha_{ref}$ from the SI-axis (z) the segment sum angle
$\alpha_{0}^{sum}$ for the backward simulation can be estimated from
a large catheter segment vector $\vec{l}_{l}$ found by distal-proximal
ray casting favoring dark lines (see step 1 below for details):

\begin{equation}
\alpha_{0}^{sum}=\arccos\left(\left\Vert \vec{r}\right\Vert \cdot\left\Vert \vec{l}_{l}\right\Vert \right).
\end{equation}

Furthermore, an estimate of the catheter length $a$ disregarding
bending using distance from the base template plane $P$

\begin{equation}
a=d_{P}\left(\vec{t}\right)
\end{equation}

is needed, where $d_{P}$ calculates point distance from a plane.

For the remaining estimation of $F_{0}$ the deflection

\begin{equation}
d=a\cdot\frac{1}{\cos\left(\alpha_{ref}\right)}\cdot\sin\left(\arccos\left(\alpha_{0}^{sum}\right)\right)
\end{equation}

from the tip to the reference catheter $\vec{r}$ in normal direction
is estimated.

With these parameter estimates, we can compute the remaining model
parameter $F_{0}^{est}$ for backward tracking of the catheter model,
i.e. a look up in a finely interpolated model table parametrized by
$a$ and $d$ yields $F_{0}^{est}$. Note, every position between
the undeflected and the most deflected catheter has a force value
$F_{intp}$ assigned that can be used as $F_{0}^{est}$. The refined
model lookup table calculates Fig.~\ref{fig:Model-catheters} nearly
without gaps (100 sample points in each direction). Linear interpolation
is used for positions still inbetween supporting $(a,d)$-points.

\subsection*{Model Guided Catheter Search Details:}

~

\begin{wrapfigure}{l}{.17\textwidth}  
	\vspace{-155pt}
	\includegraphics[bb=0 0 213 213,scale=.95]{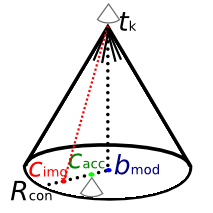}
	\vspace{-15pt} 
	\caption{\newline Search cone.}
	\vspace{-24pt}
	\label{fig:cone} 
\end{wrapfigure}


\noindent \textbf{1. Model Initializing First Cone Search} In this
phase starting from the tip $\vec{t}_{0}$ the search cone \cite{pernelle13} covers
half of the estimated catheter length $a$ and searches a first point
on the catheter approximately in the middle. A step direction vector
$\frac{a}{2}\cdot\left\Vert r\right\Vert $ 
helps in the initialization phase. The step is robust and in most
cases finds the middle of the catheter as most catheters are rather
straight from the tip up to their middle (cf. Fig. \ref{fig:Model-catheters}).
The found long segment $\vec{l}_{l}$ is the basis for the model estimation
phase described above.

\noindent Now, the main catheter detection processing starts again
from the tip $\vec{t}_{0}$, uses the initialized model as a guide,
and works its way using $N_{c}-1$ shorter segments towards the proximal
end.

\textbf{First Catheter Point Detection Iteration} ($k=0$) While the
top of the cone $\vec{t}_{0}$ is the given tip position, the base
point of the first cone search is derived as $\vec{b}_{0}=\vec{t}_{0}+\frac{a}{N_{c}-1}\cdot\left\Vert \vec{l}_{l}\right\Vert $
and the located point is accepted as $\vec{t}_{1}$. After this step,
the first short segment $\vec{l}_{s}$ is estimated and allows the
definition of the first local segment coordinate system to which the
angular spring model can relate.

\noindent \textbf{2. Definition of Local Segment Coordinate Systems}
($k=1..N_{c}-1$) With an accepted catheter segment $\vec{l}_{s}$
found by ray casting, for each segment a local right-handed coordinate
system is set-up coherent to the LPI image directions. They consist
of the catheter deflection plane normal $\vec{n}_{loc}$ defining
the 2D $\left(a,d\right)$-plane for the catheter bending simulation
(cf. Fig. \ref{fig:Model-catheters}), deflection direction $\vec{d}_{loc}$
(cf. $d$) and reference direction $\vec{r}_{loc}$ (cf. $a$). For
every iteration, they are newly defined using the last accepted segment
vector $\vec{l}_{s}=$$\vec{t}_{k}-\vec{t}_{k-1}$ ($\vec{t}_{k}$
is the top of the current cone, cf. Fig. 2)

\begin{equation}
\vec{n}_{loc}=\frac{\vec{l}_{s}}{\left\Vert \vec{l}_{s}\right\Vert }\times\frac{\vec{r}_{loc}}{\left\Vert \vec{r}_{loc}\right\Vert };\,\,\vec{d}_{loc}=\frac{\vec{n}_{loc}}{\left\Vert \vec{n}_{loc}\right\Vert }\times\frac{\vec{r}_{loc}}{\left\Vert \vec{r}_{loc}\right\Vert };\,\,\vec{r}_{loc}=\frac{\vec{n}_{loc}}{\left\Vert \vec{n}_{loc}\right\Vert }\times\frac{\vec{d}_{loc}}{\left\Vert \vec{d}_{loc}\right\Vert }.
\end{equation}

\noindent This allows the model catheter to deflect in more than one
plane and makes the algorithm truly 3D capable. From step to step,
the local coordinate systems slightly rotate. For increased efficiency,
we also do not need to use more than one spring for the rod joints
as proposed in \cite{goksel09}.

\noindent \textbf{3. Model-based Cone Search} The top point of the
cone $\vec{t}_{k}$ is a definitely accepted catheter point and the
base point defining the center of a circular region as cone base is
proposed as follows (cf. Fig. \ref{fig:cone}): Metrically synchronously
to the cone search and subject to the local coordinate system, deflection
backward simulation steps take place. Finally, a new model cone base
point $\vec{b}_{mod}$ is proposed adding a model-based step vector
to the current cone top $\vec{t}_{k}$ resulting in

\begin{equation}
\vec{b}_{mod}=\vec{t}_{k}+d_{seg}\cdot\left(\frac{\vec{d}_{loc}}{\left\Vert \vec{d}_{loc}\right\Vert }\cdot sin\left(\alpha_{i}^{sum}\right)+\frac{\vec{r}_{loc}}{\left\Vert \vec{r}_{loc}\right\Vert }\cdot cos\left(\alpha_{i}^{sum}\right)\right).
\end{equation}

\noindent Here, $d_{seg}=a/\left(N_{c}-1\right)$ represents the chosen
segment length. If the point $\vec{c}_{img}$ found by ray casting
inside the search cone at the base fulfills the distance constraint

\begin{equation}
\left\Vert \vec{c}_{img}-\vec{b}_{mod}\right\Vert <d_{tol}
\end{equation}

it is accepted valid as next cone top $\vec{t}_{k+1}$ and added to
the list of catheter defining points (see Fig. \ref{fig:cone}).

\textbf{Model Constraint Violation} Otherwise, if the point $\vec{c}_{img}$
found from image features deviates too far from the model proposed
base point $\vec{b}_{mod}$ an acceptable compromise point $\vec{c}_{acc}$
is generated and inserted into the list of found catheter points:

\begin{equation}
\vec{c}_{acc}=\vec{b}_{mod}+\frac{\vec{c}_{img}-\vec{b}_{mod}}{\left\Vert \vec{c}_{img}-\vec{b}_{mod}\right\Vert }\cdot min\left(d_{tol},\frac{\left\Vert \vec{c}_{img}-\vec{b}_{mod}\right\Vert }{2}\right).
\end{equation}

By this means, the resulting catheter trajectory is a compromise or
hybrid between image feature based and model based point proposals.

\textbf{Image Features Preferred and Final Catheter Definition} Catheter
trajectories appear as signal voids in MR scans, while the surrounding
tissue is brighter. Thus, we use a small circular gradient inspired
2D mask that responds to this pattern and also consider the expected
geometry of the catheter cross-sections (diameter). When searching
in conic regions by ray casting, we look for distal to proximal line
integrals minimizing the image intensity value taking the preferred
gradient pattern into account by subtracting it from the sum. Finally, the hybrid model and image feature based
points are approximated by a Bezier curve \cite{pernelle13} with $N_{c}$ control points.

\section{Experiments}

For evaluation we used MR images from ten patients in which 101 catheters
were carefully segmented by a medical expert. The MRI scans were acquired
using a mix of protocols that include two-dimensional low-artifact
T2-weighted Fast Spin Echo (FSE, TR/TE=3000/120 ms, 0.2x0.3x2.0 mm$^{3}$),
three-dimensional FSE (Siemens SPACE, TR/TE=3000/160 ms, 0.4x0.4x1.0
mm$^{3}$), 3D balanced steady state free precession (3D bSSFP, TR/TE
= 5.8/2.9 msec, 0.6x0.6x1.6 mm$^{3}$). The scanner was a 3 Tesla
MRI ``Magnetom Verio'' (Siemens Healthcare, Erlangen, Germany).
The catheters in the images appear as signal voids (dark trajectories)
on the T2/FSE images, and are surrounded by prominent so called blooming
artifacts in the 3D bSSFP sequence.

The plastic catheters (diameter 1.6 mm or 16 Gauge) have a metallic
needle inlay made of tungsten-alloy and an average insertion depth
of 74 mm. We use 7 search cones with equal height $d_{seg}=a/\left(N_{c}-1\right)$,
$N_{c}$=8. Our synchronous catheter model simulation always uses
$N_{seg}=20$ segments for full length (187 mm). We empirically
choose $k_{a}=2050\,\mu$Nm as our 16 Gauge catheters are stiffer
than those (18 Gauge) presented in \cite{goksel09}. The radius of
the cone base circle $R_{con}$ is set generously to 20 mm to evaluate
the model vs. the image features influence.

Three experiments are conducted in this work by varying the tolerance
$d_{tol}$ as 0/$\infty/1$ mm: (1) ``full model'' using guidance
model points only, (2) ``image features only'' with fully image
feature based points and (3) ``trade-off'' as a compromise or hybrid
mode between model and image features based points as core contribution
of this paper.

\section{Results}

The Hausdorff distance (HD) between the gold standard and automatic
segmentations in the patient MR images was evaluated. The HD quantifies
the maximal distance between two catheter tubes \cite{huttenlocher93}
and thus is a very strict measure which is necessary for evaluation
of clinical applicability.

\begin{figure}[t]
\centering{}\includegraphics[scale=0.2,bb=0 0 553 518]{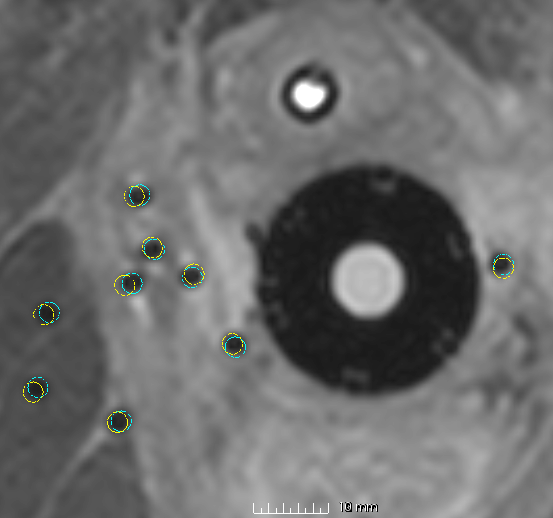}\,\includegraphics[scale=0.2,bb=0 0 553 517]{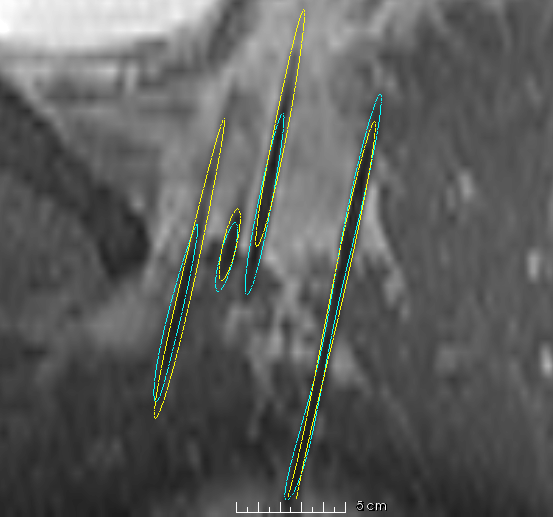}\,\includegraphics[bb=143bp 0bp 393bp 518bp,clip,scale=0.2]{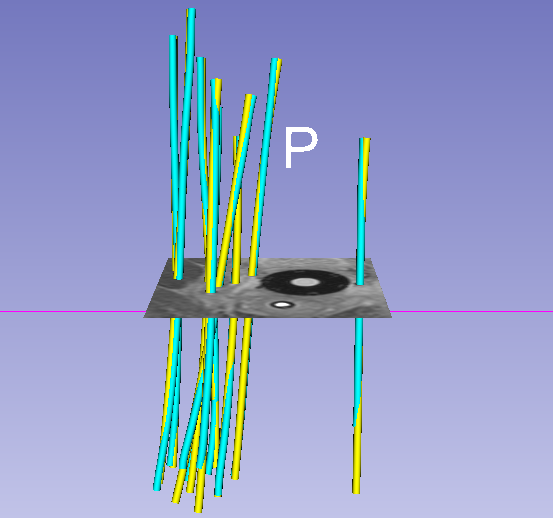}\pagebreak{}

\includegraphics[scale=0.2,bb=0 0 521 518]{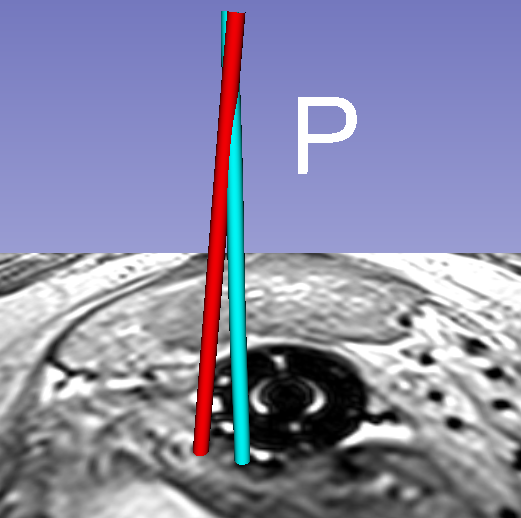}\,\includegraphics[scale=0.2,bb=0 0 521 518]{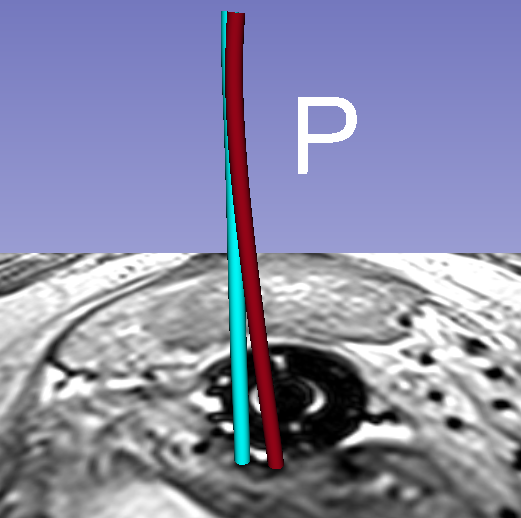}\,\includegraphics[scale=0.2,bb=0 0 521 518]{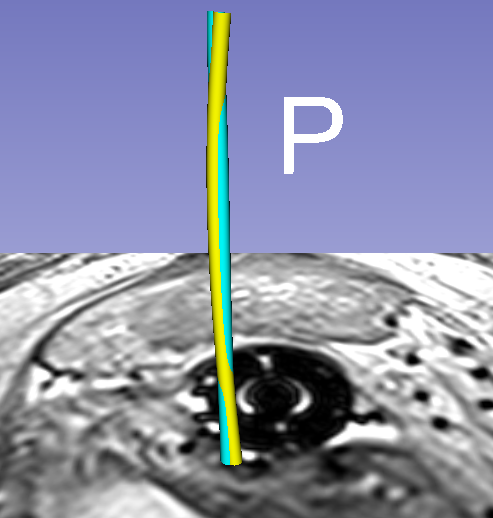}\protect\vspace{-0.44cm}
 \protect\protect\protect\caption{\label{fig:Case-study}Examples of two cases. Top row: (left) axial
view, (middle) sagittal view, (right) 3D rendering with axial plane.
Automated segmentations of catheters (yellow) from a T2-weighted MR
scan of a gynecologic brachytherapy patient are shown in yellow and
manual segmentations in cyan. Bottom row: Segmentation of a particular
catheter that fails when the method ignores image features (left,
red), fails when it ignores the model (middle, dark red), and succeeds
when it incorporates both model and image features (right, yellow). }
\end{figure}

\begin{figure}[!h]
\noindent \raggedright{}%
\begin{minipage}[c]{0.2\textwidth} \protect\vspace{-3.7cm}
\includegraphics[scale=0.36,bb=0 0 2075 460]{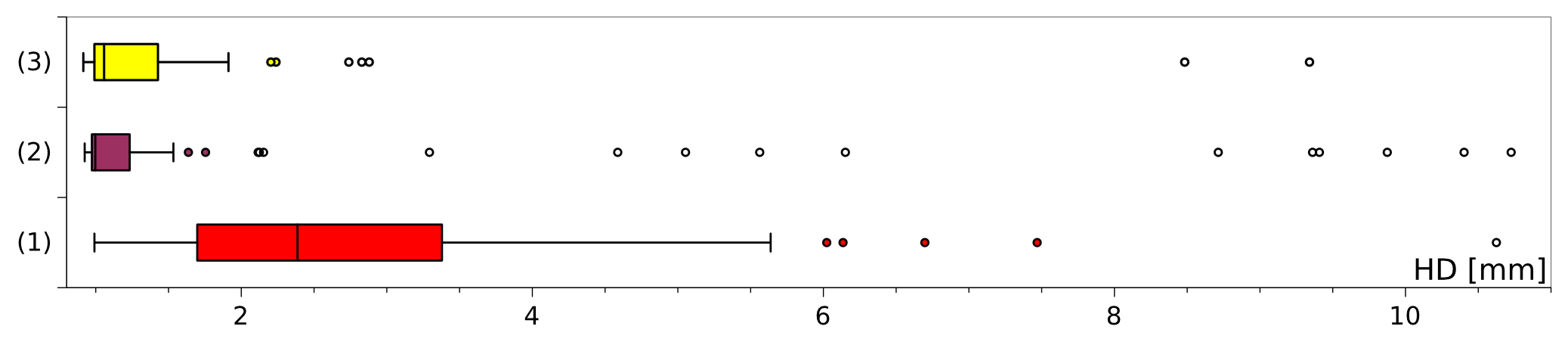}%
\end{minipage}\hspace*{7.1cm}%
\begin{minipage}[c]{0.2\textwidth}%
{\scriptsize{}{}{}{}}%
\begin{tabular}{|c|c|c|}
\hline 
{\scriptsize{}{}{}{}CO}  & {\scriptsize{}{}{}{}Med.}  & {\scriptsize{}{}{}{}Mean$\pm$StD}\tabularnewline
\hline 
{\scriptsize{}{}{}{}\cellcolor{Yellow}}\textbf{\textcolor{black}{\scriptsize{}{}{}2}}{\scriptsize{}{}
}  & {\scriptsize{}{}{}{}\cellcolor{Yellow}}\textbf{\textcolor{black}{\scriptsize{}{}{}1.05}}{\scriptsize{}{}
}  & {\scriptsize{}{}{}{}1.40$\pm$1.15}\tabularnewline
\hline 
{\scriptsize{}{}{}{}\cellcolor{Wine}}\textbf{\textcolor{white}{\scriptsize{}{}{}12}}{\scriptsize{}{}
}  & {\scriptsize{}{}{}{}\cellcolor{Wine}}\textbf{\textcolor{white}{\scriptsize{}{}{}1.00}}{\scriptsize{}{}
}  & {\scriptsize{}{}{}{}1.99$\pm$2.79}\tabularnewline
\hline 
{\scriptsize{}{}{}{}\cellcolor{Red}}\textbf{\textcolor{white}{\scriptsize{}{}{}37}}{\scriptsize{}{}
}  & {\scriptsize{}{}{}{}\cellcolor{Red}}\textbf{\textcolor{white}{\scriptsize{}{}{}2.39}}{\scriptsize{}{}
}  & {\scriptsize{}{}{}{}2.75$\pm$1.54}\tabularnewline
\hline 
\multicolumn{3}{c}{\vspace{-0.1cm}
 }\tabularnewline
\end{tabular}%
\end{minipage}\protect\vspace{-0.44cm}
 \protect\protect\protect\caption{\label{fig:Results}Result boxplots (left) and statistics as legend
(right) of the three experiments: Boxplots with Hausdorff distances
(HD) {[}mm{]} and critical outliers (CO, HD$>$3 mm). For experiment
(2) one outlier at 18 mm is not shown in the graphics for display
reasons; or overlap or are located in the maximum range (whisker).}
\end{figure}

Qualitative results for two different examples can be seen in Fig.
\ref{fig:Case-study} and demo movies\footnote{https://goo.gl/fLsa2R}. The top rows illustrates cross-sectional and
3D rendered views of automated segmentations of a catheter array.
The overlap of the colored catheters shows how well they agree. The
bottom row of this figure illustrates the case of one particular catheter
from the experiments, and specifically how both the model and image guidance are necessary to achieve correct segmentation. 

Quantitatively, as can be seen in Fig. \ref{fig:Results}, comparing
experiment (2) and (3) the number of critical outliers (HD$>$3 mm)
can be significantly reduced to one sixth. Outliers
with $>$2 mm HD can be reduced from 15 to 7 (53\%). For (3), the threshold
for the number of outliers to increase from 2 to 3 is 2.88 mm. Regarding
(3), the median value that represents the most frequent result in
a clinical setting is not significantly higher than for using image
features alone (2). Apart from that, using the estimated model alone
is not advisable, as the mean/median values as well as the indicators
of robustness (standard deviation, quartile ranges) are too high.
This is reflected in the statistics: The differences between the results
of the experiments are statistically different (Kruskal-Wallis ANOVA
on ranks test: $p<0.001$). Tukey post-hoc tests show differences
between experiment (1) vs. (2) and (3) to be significant ($p<0.05$),
while (2) and (3) do not differ. The segmentation time for a single
catheter is 2 seconds on an standard Intel i7 3 GHz computer (Python implementation in Slicer\footnote{http://www.slicer.org/}).

\section{Conclusion}
The incorporation of our proposed model strongly helps to shift
the state of the art in catheter segmentation from MRI \cite{pernelle13} towards practical
clinically usability by reducing outliers to one-fourth (there 8 needles with HD$>$4 mm).

The proposed model guided catheter segmentation approach (exp. (3)) can reduce
the number of outliers with HD$>$3 mm to one sixth, compared to using image features alone (exp. (2)). The number of
outliers with HD$>$2 mm can be reduced by half between these two experiments.  The benefit of the model is to filter out implausible
outliers from the image-based feature points which would cause improbable
curvatures in the segmented catheter trajectory. This is an important
contribution towards enabling practical clinical acceptance of MRI-guided interventions,
a trend that is gaining momentum in recent years.

\section{Discussion and Future Work}

\noindent We have shown that a hybrid strategy is necessary for catheter
segmentation from MRI because alone, neither image feature nor model
based catheter point estimations, delivers consistently correct results.
The initialization of the model is an area that needs further investigation,
especially regarding more bent cath-eters. 
 \bibliographystyle{plain}

\end{document}